%% file: main.tex
\documentclass{article}
\usepackage{spconf,amsmath,graphicx}

\usepackage[normalem]{ulem}
\usepackage{booktabs}

\usepackage{amssymb}% http://ctan.org/pkg/amssymb
\usepackage{pifont}% http://ctan.org/pkg/pifont
\newcommand{\cmark}{\ding{51}}%
\newcommand{\xmark}{\ding{55}}%

\title{Towards Controlled Table-to-Text Generation \\ with Scientific Reasoning}

\name{Zhixin Guo$^{1}$, Jianping Zhou$^{1}$, Jiexing Qi$^{1}$, Mingxuan Yan$^{1}$, Ziwei He$^{1}$, Xinbing Wang$^{1}$, Chenghu Zhou$^{2}$}

\address{$^{1}$Shanghai Jiao Tong University, Shanghai, China \\
$^{2}$ IGSNRR, Chinese Academy of Sciences, Beijing, China}

\begin{document}
\ninept
\maketitle
\input{Files/abstract.tex}
\input{Files/introduction.tex}

\input{Files/dataset.tex}

\input{Files/task_def.tex}

\input{Files/baselines.tex}

\input{Files/experiments.tex}

\input{Files/conclusion.tex}

\bibliographystyle{IEEEbib}
\bibliography{mybib}

\end{document}

%% file: Files/abstract.tex
\begin{abstract}
The sheer volume of scientific experimental results and complex technical statements, often presented in tabular formats, presents a formidable barrier to individuals acquiring preferred information. The realms of scientific reasoning and content generation that adhere to user preferences encounter distinct challenges. In this work, we present a new task for generating fluent and logical descriptions that match user preferences over scientific tabular data, aiming to automate scientific document analysis. To facilitate research in this direction, we construct a new challenging dataset CTRLSciTab consisting of table-description pairs extracted from the scientific literature, with highlighted cells and corresponding domain-specific knowledge base. We evaluated popular pre-trained language models to establish a baseline and proposed a novel architecture outperforming competing approaches. The results showed that large models struggle to produce accurate content that aligns with user preferences. As the first of its kind, our work should motivate further research in scientific domains.
\end{abstract}

%% file: Files/introduction.tex
\section{Introduction}
Table-to-text generation is a significant research problem involving the generation of analytical descriptions from tabular data. Recently, pre-trained language model (PLM)-based approaches have demonstrated impressive improvements in text fluency and fidelity compared to traditional template-based methods, as seen in the popular table-to-text challenges, WikiBio \cite{liu2018table}, LogicNLG \cite{RN153}, RottoWire \cite{wiseman2017challenges}, and ToTTo \cite{RN115}. However, their success is highly dependent on pre-training using a large corpus of open-domain text, such as the two billion words of Wikipedia \cite{denoyer2006wikipedia}. 

\begin{figure}
\centerline{\includegraphics[scale=0.37]{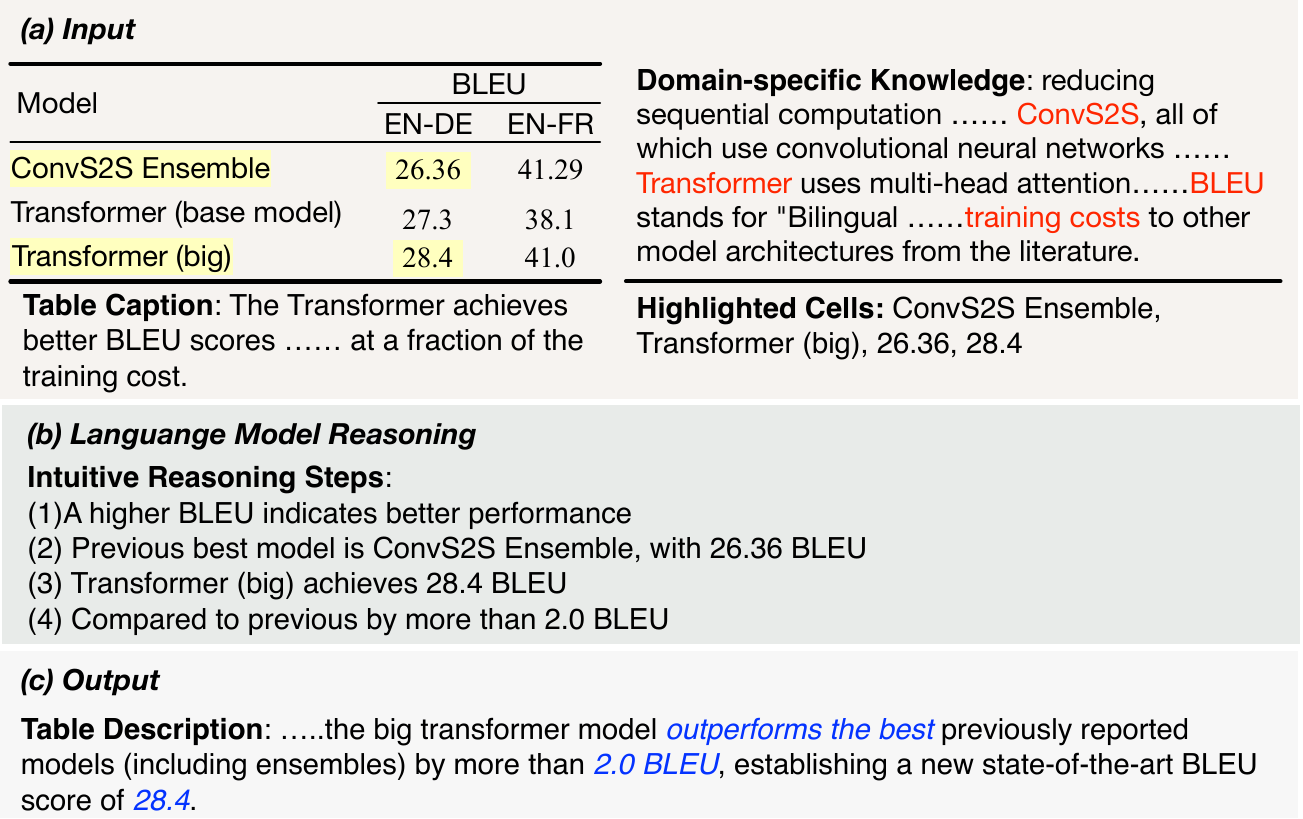}}
\caption{An illustration of controlled table-to-text generation incorporating explicit scientific reasoning stages. (a) represents the input information, (b) illustrates the inherent reasoning processes of language models, and (c) displays the resultant descriptions. Yellow highlights user preferences; red relates to tabular knowledge, and blue indicates scientific reasoning content. Potential scientific reasoning steps are outlined at the bottom. The original table is adapted from \cite{vaswani2017attention}.}
    \label{tablesample}
\end{figure}

Scientific domains that require advanced expertise present a significant challenge for PLM-based NLG systems in generating descriptions with scientific reasoning. Scientific reasoning is a cognitive process characterized by systematic, logical, and evidence-based reasoning to understand the natural world, develop hypotheses, design experiments, evaluate evidence, and draw conclusions based on empirical data and observations \cite{schunn1999generality}, \cite{bao2009learning}, \cite{zimmerman2000development}. In the context of scientific natural language processing (NLP), scientific reasoning represents a significant bottleneck, relying primarily on domain-specific knowledge tailored to specific scientific phenomena. 

While existing challenges such as SciGen \cite{moosavi2021scigen} and numericNLG \cite{suadaa2021towards} address table-to-text generation related to scientific content, they predominantly focus on numerical reasoning. We argue that scientific reasoning in the context of scientific NLP presents a unique and comprehensive challenge for table-to-text generation. Moreover, with the increasing use of PLM-based natural language generation (NLG) systems in real-world applications, the focus has shifted from the generation of generic content to the generation of customized content that is aligned with user preferences. However, the currently popular methods of table-to-text tasks may not consistently match user preferences due to the broad nature of the generated output.

In this work, we propose a new task of controlled table-to-text generation with scientific reasoning, which aims to generate analytical descriptions that are consistent with domain-specific knowledge and align with user preferences. To facilitate research on this task, we present CTRLSciTab, which consists of 8,967 scientific table-description pairs with an external domain-specific knowledge base and highlighted cells. CTRLSciTab represents a challenging table-to-text generation task with two unique features: (1) all pairs are from the scientific literature, requiring scientific reasoning, and (2) all descriptions are aligned with user preferences. 

We conducted extensive experiments on popular PLM-based models, revealing their poor performance in scientific domains requiring advanced expertise, a persistent challenge for the NLP community. To address this, we propose a retrieval-based pre-trained model trained on our dataset, which shows a better performance. However, evaluations, both automatic and human, indicate that the generated sentences may sometimes contain incorrect and hallucinatory content, highlighting this data set's potential as a valuable benchmark for evaluating controlled table-to-text generation with scientific reasoning.

%% file: Files/dataset.tex
\section{The CTRLSciTab Dataset}
\begin{figure}
\centerline{\includegraphics[scale=0.34]{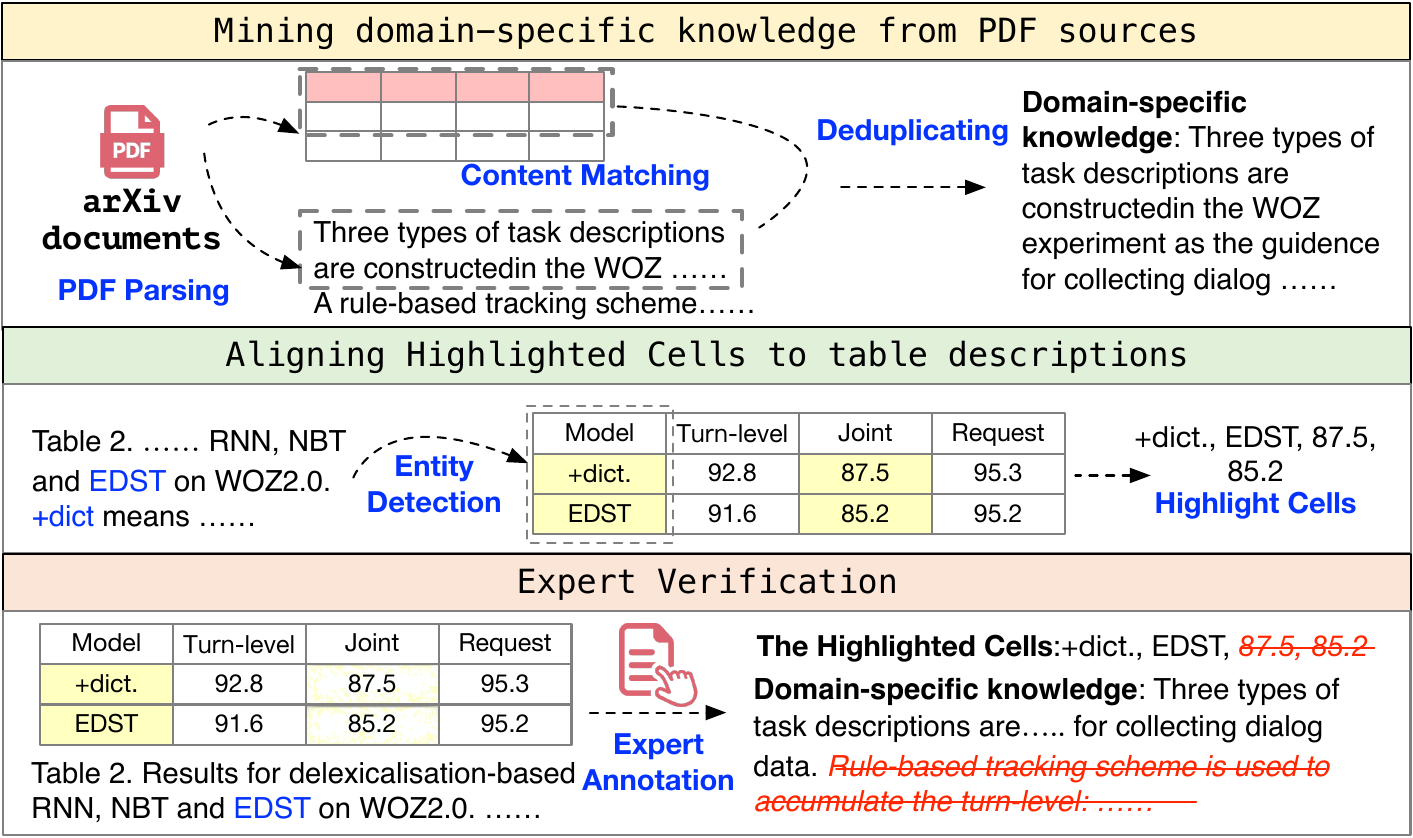}}
    \caption{Overview of CTRLSciTab construction steps, including mining domain-specific knowledge from PDF source, aligning highlighted cells to table descriptions, and expert veriﬁcation.}
    \label{annotation}
\end{figure}

\textbf{Data Preparation}. CTRLSciTab, a broad dataset, consolidates table-to-text pairs from the \cite{moosavi2021scigen} and \cite{suadaa2021towards} datasets, which originally obtained these pairs from scientific literature with a focus on numerical reasoning. To cater to controlled generation coupled with scientific reasoning, we embarked on a retrieval process for original scientific articles from arXiv.org using tabular metadata. Subsequently, we instituted a procedure to extract sentences associated with tables, thereby constructing a domain-specific knowledge base for scientific reasoning. We also highlighted cells mentioned within table descriptions, serving as prompts instructing PLMs to mirror user preferences. As depicted in Figure~\ref{annotation}, the extraction we first transform articles into XML-encoded files. A greedy algorithm is then engaged for content matching, aligning parsed content with tables, and preserving entity-referencing sentences. Deduplication removes any overlap with table descriptions, and entity detection identifies pertinent entities within the descriptions.

\textbf{Annotation Procedure}. Potential inaccuracies in the annotated data, such as irrelevant highlighted cells, are rectified to ensure quality. Expert annotators, who are computer science undergraduates, verify the automatically annotated data. The tasks involve refining domain-specific knowledge and annotating highlighted cells. Annotators eliminate sentences unrelated to tabular data or descriptions, retaining content explicitly stated or logically inferred from the data or descriptions.  To quantify the annotation agreement, we randomly selected 100 samples, and two expert annotators achieved a 66.7\% agreement rate on the highlighted cells and a 70.6\% agreement rate on domain-specific knowledge, thereby underscoring the necessity of domain-specific knowledge.

\textbf{Data Analysis}. CTRLSciTab is a large data set that contains 8,967 controlled table-description pairs with rich domain-specific knowledge in various scientific domains. Each table in CTRLSciTab consists of 52 cells on average, with a description of 34 words and highlighted cells as controlled preferences which occupy about 20\% of the cells. To support common sense generation, we construct an average of 20 domain-specific knowledge sentences per table. 

% Figure~\ref{tablestatics} demonstrates the statistics for the top ten domains of CTRLSciTab. The rich and diverse scientific domains covered in CTRLSciTab make it an ideal benchmark for evaluating the performance of models in generating controlled natural language descriptions from scientific tabular data with common sense reasoning.

In the context of table-to-text generation datasets, CTRLSciTab introduces a distinct perspective on content selection and surface realization that distinguishes it from other existing datasets. Prior to the advent of neural approaches, generation systems typically separated content selection (what to say) from surface realization (how to say it) \cite{reiter1997building}. Traditional generation datasets focused predominantly on a single phase, simplifying the overarching complexity of the task. ToTTo \cite{RN115} was seminal in presenting a controlled table-to-text generation task with an emphasis on content selection. However, its design does not adequately address the needs of scientific scenarios that require expertise. Some recent datasets have proposed incorporating complex reasoning, such as numerical \cite{suadaa2021towards}, \cite{moosavi2021scigen} and logical reasoning \cite{RN153}, into the task by framing it as a summarization problem focused on surface realization. A deeper exploration of the chain of thought \cite{wei2022chain} in Large Language Models (LLMs) has led to remarkable progress in such generation tasks. CTRLSciTab uniquely challenges generation systems by imposing controlled constraints on content selection while requiring sophisticated scientific reasoning for surface realization.

\begin{table}
\begin{center}

\resizebox{\columnwidth}{!}{
  \begin{tabular}{lccccccc} 
\toprule
\textbf{Data set} & \textbf{Pairs} & \textbf{Cell}  & \textbf{Domain}  & \textbf{Scientific Reasoning} & \textbf{Controlled}   \\ \hline
WikiBIO & 400K & 17&  Open & \xmark & \xmark \\ 
ToTTo  & 136K & 3 &  Open & \xmark & \cmark  \\  
LogicNLG  & 37K & 91&  Open & \xmark  & \xmark \\ 
SciGen & 53K  & 55 &  Scientiﬁc  & \xmark & \xmark \\ 
numericNLG & 1.3K & 46  &  Scientific  & \xmark & \xmark \\ 
CTRLSciTab & 9.0K & 52 &  Scientific & \cmark & \cmark  \\ \bottomrule
\end{tabular}
}
\end{center}
{
\caption{Comparison of CTRLSciTab and previous table-to-text generation data sets. The pairs represent the number of annotated structure data in each data set. Cell denotes the average number of total cells.}
\label{table_overview}
}
\end{table}

% \begin{figure}
% \centerline{\includegraphics[scale=0.28]{Images/Statistics.pdf}}
% \caption{Statistics for the top ten domains of CTRLSciTab}
%     \label{tablestatics}
% \end{figure}

%% file: Files/task_def.tex
\section{Task Definition}
\begin{figure*}[!t]
\centering
\includegraphics[scale=0.36]{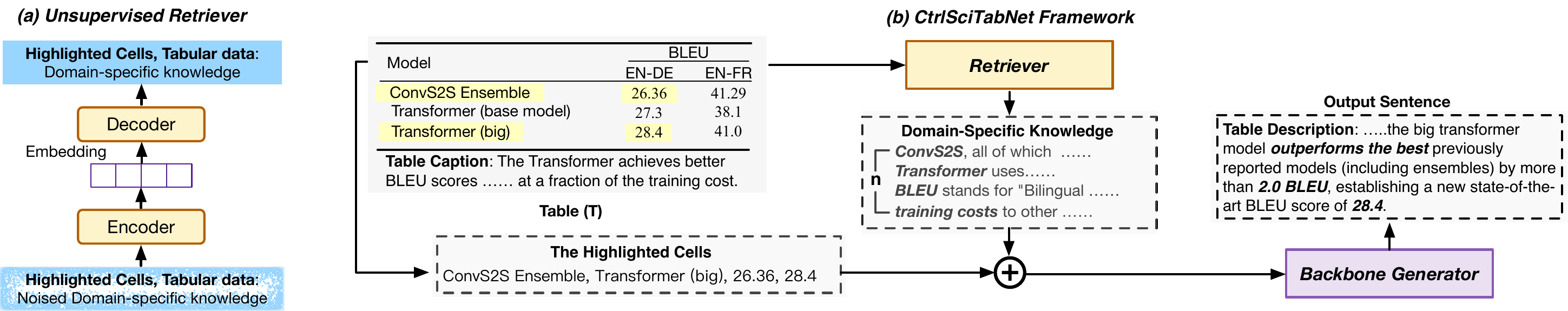}
\caption{An illustration of the CTRLSciTabNet structure: (a) depicts the architecture of our unsupervised retriever; (b) outlines the two-step operation of CTRLSciTabNet, which involves a retriever selecting the top-$n$ related domain-relevant sentences, followed by a pre-trained language model, the generator, utilizing this data alongside tabular inputs and highlighted cells.}
\label{framework}
\end{figure*}

This task aims to generate natural language descriptions that are both fluent and accurate, incorporating domain-specific knowledge while remaining consistent with the tabular data and user preferences. The input consists of structured data, highlighted cells, and domain-specific knowledge, denoted as $D={(T, H, B)}$. Here, $T$ signifies a linearized table, with $T = \left\{t_1,\cdots, t_{|n|}\right\}$. Each tabular data, $t_i$, consists of an attribute-value pair, where $a_i$ and $v_i$ can take values such as strings, numbers, phrases, or sentences. The highlighted cells are akin to $t_i$ and are denoted by $H = \left\{h_1,\cdots, h_{|n|}\right\}$, acting as prompts reflecting user preferences. Furthermore, $B = \left\{b_i,\cdots,b_{|m|}\right\}$ represents domain-specific knowledge, with each $b_i$ corresponding to a sentence associated with the tabular data. The expected output is an analytical description aligned with user preferences and incorporating domain-specific knowledge $R$. 

% Fluency, measured on a five-point Likert scale, assesses the grammatical correctness and naturalness of sentences. Faithfulness measures sentence accuracy against table data, with annotators determining the proportion of faithful content. Recall assesses sentence relevance to highlighted cells, with all sentence parts aligned with cells deemed relevant. Annotators determine the proportion of such facts within the generated descriptions. Finally, Valid Facts determines the proportion of unique and accurate content within generated sentences relative to reference sentences.

%% file: Files/baselines.tex
\section{Methods}

The abundance of supporting domain-specific knowledge for each table, with an average of over 20 sentences, makes it challenging to leverage using popular PLMs. To address this, we propose a retriever-generator framework, CTRLSciTabNet, as our initial approach to the problem. The overall framework of our method is illustrated in Figure~\ref{framework}. The retriever retrieves supporting domain-specific knowledge based on the table contents, which is then used by the generator to produce descriptions according to the highlighted cells.

\textbf{Retriever}. In this study, as delineated in Figure~\ref{framework} (a), we introduce an unsupervised sentence embedding technique under controlled conditions. Drawing inspiration from \cite{wang2021tsdae}, our approach seeks to derive embeddings of domain-specific sentences represented by $B$ through the process of reconstructing them from a perturbed corpus, denoted as $\tilde{B}$. Further refining the embedding process, we impose constraints$H$ and $T$, signifying user preferences and the original tabular data, respectively. The conditional distribution $P(\tilde{B}|B, T, H)$. The training goal of  the retriever is defined by Equation~\ref{equ1}:
\begin{equation}
\begin{split}
    J_{SDAE}(\theta) = E_{x\sim D} [\log {P_{\theta}(x|\tilde{x}})] \\
    = E_{x\sim D} [\sum_{t=1}^{L} \log {P_{\theta}(x|\tilde{x}})] \\
    = E_{x\sim [B : T : H]} [\sum_{t=1}^{l_{|B|}+l_{|T|}+l_{|H|}} \log {\frac{exp(h_{t}^{T}e_{t})}{\sum_{i=1}^{N}exp(h_{t}^{T}e_{i})}})]
\end{split}
\label{equ1}
\end{equation}
where $L = l_{|B|}+l_{|T|}++l_{|H|}$ denotes the total length of input tokens, including the highlighted cells, table contents, and domain-specific knowledge. 

\textbf{Generator}. We aim to develop a sentence generator that can produce coherent and sensible descriptions based on tabular data and the highlighted cells while incorporating domain-specific knowledge according to user preferences. To achieve this goal, we employ BART-base \cite{lewis2020bart} and T5-small \cite{raffel2020exploring} for the sentence generator. The learning objective of our sentence generator is to minimize the cross-entropy loss. Specifically, given the tabular data $T$, highlighted cells $H$, domain-specific knowledge $B$, and corresponding table descriptions $R$, the learning objective is defined as Equation~\ref{equ2}:
\begin{equation}
    L_{LM} = - \sum_{i=1}^{|R|} \log{P_{G}}{(R_{i}|R_{<i}; E([H:T:B]))}
\label{equ2}
\end{equation}
where E represents the encoder.

%% file: Files/experiments.tex
\section{Experiments}
\begin{table}
\centering
\resizebox{\columnwidth}{!}{
\begin{tabular}{lcccc} \toprule
\textbf{Baselines }         & \textbf{ BLEU$\uparrow $} & \textbf{ METEOR$\uparrow $} & \textbf{ BertScore$\uparrow $} & \textbf{ BLEURT$\uparrow $} \\ \hline
TF-IDF + DG (Bart)  & 1.60                      & 0.09                        & 0.75                       & -0.93                       \\
Retriever + DG (Bart)                  & 2.00                      & 0.09                        & 0.78                       & -1.00                       \\
Retriever + DG (GPT-3.5) & 4.76 & 0.20 & 0.84 & -0.51 \\
CTRLSciTabNet (Bart)               & \textbf{\textbf{16.90 }}  & \textbf{\textbf{0.34 }}     & \textbf{\textbf{0.87}}    & \textbf{\textbf{-0.32 }}    \\
CTRLSciTabNet (T5)              & 6.60                      & 0.29                        & \uline{0.85}                       & -0.68                       \\ \hline
DG w/o BKG (GPT-3.5) & 3.07 & 0.17 & 0.82 & -0.65 \\
Bart w/o BKG               & \uline{14.90 }            & \uline{0.31 }               & \textbf{\textbf{0.87}}    & \uline{-0.38 }              \\
T5 w/o BKG               & 2.10                      & 0.18                        & 0.82                       & -0.87                       \\
 \bottomrule
\end{tabular}}
\caption{Automatic evaluation results of all methods. The symbol $\uparrow$ indicates that a higher score represents better performance. The best result is \textbf{bold}, and the second best is \underline{underlined} within each metric. W/o BKG denotes the model without using domain-specific knowledge. DG denotes direct generate from the pre-trained model without fine-tuning}
\label{table4}

\end{table}

% \subsection{Other Baseline Models}
\textbf{Baselines}. To evaluate the effectiveness of our approach, we compare it to several popular strategies by replacing the retriever and sequence generator as baselines. Specifically, we consider the following methods:

\textit{TF-IDF + Direct Generation (Bart-base)}. We implement TF-IDF \cite{aizawa2003information}, a commonly utilized retrieval method, to access domain-specific knowledge. The architectural design remains consistent with our framework, utilizing the Bart-base model without fine-tuning for the generative tasks.

\textit{Retriever + Direct Generation (Bart-base)}. To assess the efficacy of our proposed retrieval method, we substitute the TF-IDF retriever with our approach, maintaining all other settings.

\textit{Retriever + Direct Generation (GPT-3.5)}. We conduct an additional investigation using our dataset to further explore the functionality of GPT-3.5 \cite{brown2020language}.

\label{section:auto_metrics}
\textbf{Automatic Evaluation}. To evaluate the performance of our model, we use ﬁve automatic metrics: BLEU [7], METEOR [8], BertScore [9], and BLEURT [10]. BLEU and METEOR are widely used metrics that measure the informativeness of generated text. BertScore and BLEURT are pre-trained metrics to measure the similarity between the generated descriptions and the reference sentence. Table~\ref{table4} delivers a thorough assessment of all tested systems using metrics defined in $\S$\ref{section:auto_metrics}. To evaluate the retrievers' performance, we maintain our model's architecture and retrieve the top-3 related domain-specific knowledge sentences using our proposed and TF-IDF-based retrievers. We then generate the corresponding descriptions directly from the retrieved sentences. Our proposed retriever outperforms the TF-IDF-based retriever based on the automatic evaluation metrics. Subsequently, we use the same retriever results as input for all generator models, which allows us to evaluate the performance of the generator part. It is worth noting that the currently popular large-scale GPT-3.5 still does not perform well on tasks that require domain-specific knowledge for scientific reasoning.

\begin{table}
\centering
\resizebox{\columnwidth}{!}{
\begin{tabular}{lcccc} \toprule
\textbf{Model}    & \textbf{Fluency$\uparrow$} & \textbf{Faithfulness$\uparrow$} & \textbf{Recall$\uparrow$} & \textbf{Valid Facts$\uparrow$} \\ \hline
Bart w BKG    & \textbf{4.19 }             & \textbf{0.39 }                  & \textbf{0.40 }            & \textbf{0.88 }                 \\
Bart w/o BKG     & \uline{4.14 }              & 0.31                            & \uline{0.38 }             & \uline{0.80 }                  \\
T5 w BKG    & 3.74                       & \uline{0.33 }                   & 0.33                      & 0.50                           \\
T5 w/o BKG     &  2.60                      &  0.30                           &  0.24                     &  0.47                          \\
 \bottomrule
\end{tabular}
}
\caption{Performance of human evaluation result of CTRLSciTabNet using the BART-base and T5-small models based on several metrics, including fluency, faithfulness, recall of covered highlighted cells, and valid factors. $\uparrow$ denotes a higher score representing a better performance. The best models are \textbf{bold}, and the second best ones are \underline{underlined} within each evaluation. W/o BKG denotes the model without the use of domain-specific knowledge.}
\label{human_evaluation}
\end{table}

In addition, we provide a breakdown of the performance of using domain-specific knowledge. When we use only the tabular data as input, the performance for all PLMs (GPT-3.5, Bart-base, and T5-small) significantly lags behind the results obtained when external domain-specific knowledge is incorporated. These results demonstrate the importance and necessity of domain-specific knowledge as expertise for scientific reasoning. 

Except for BertScore, the performance of all baseline systems is significantly poor in the remaining metrics. Our human evaluation in $\S$\ref{section:human_evaluation} reveals that the majority of the generated descriptions are of poor quality. Additionally, existing automatic evaluation metrics fail to evaluate the coherence of the highlighted cells and the generated descriptions for controlled table-to-text generation. Furthermore, CTRLSciTabNet, based on the Bart-base model, outperforms the other models on most of the automatic evaluation metrics. These results highlight the inadequacy of automatic evaluation metrics for this challenge.

\label{section:human_evaluation}
\textbf{Human Evaluation}.
Existing automatic metrics have difﬁculty assessing the coherence between highlighted cells and generated sentences, requiring human evaluation using four different criteria: ﬂuency, ﬁdelity, recall, and valid facts. Fluency, measured on a ﬁve-point Likert scale, assesses the grammatical correctness and naturalness of sentences. Faithfulness measures sentence accuracy against table data, with annotators determining the proportion of faithful content. Recall assesses sentence relevance to highlighted cells, with all sentence parts aligned with cells deemed relevant. Annotators determine the proportion of such facts within the generated descriptions. Finally, Valid Facts determines the proportion of unique and accurate content within generated sentences relative to reference sentences.

Table \ref{human_evaluation} shows the results of human evaluation, highlighting the superior performance of models using external domain-specific knowledge in terms of fluency, fidelity, recall, and valid facts. The data suggest that advanced domain knowledge is essential for producing scientifically sound results in the scientific domain. Our proposed method, CTRLSciTabNet (Bart-base w BKG), consistently outperforms all evaluated metrics, in line with the results of automatic evaluation metrics. 

However, the poor fidelity score indicates a significant gap within the CTRLSciTab dataset that needs to be addressed. In addition, the poor recall of highlighted cells implies that the evaluated models fall short in performing controlled generation independent of scientific reasoning. Both automatic and human evaluation results support CTRLSciTab as a practical testbed for controlled table-to-text generation with scientific reasoning. In particular, CTRLSciTabNet (Bart-base w BKG) shows the best performance across all metrics (sign test with a p-value \textless{0.05}).

\begin{figure}
\centerline{\includegraphics[scale=0.39]{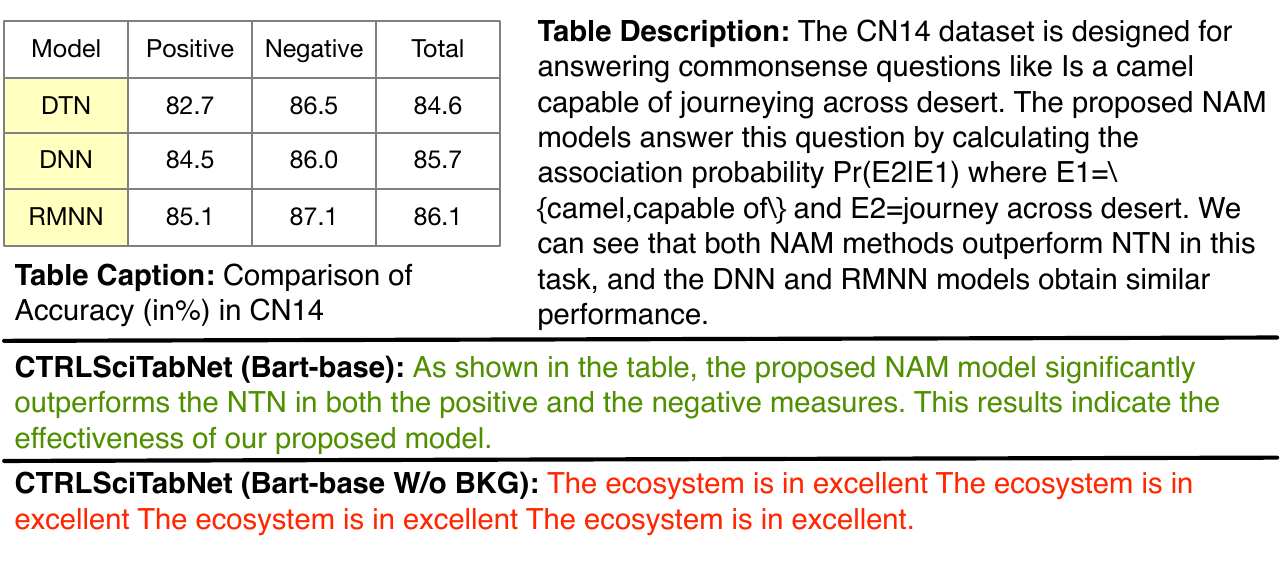}}
\caption{Case study of CTRLSciTabNet. Contents in yellow cells indicate the highlighted cells. W/o BKG denotes the model without the use of domain-specific knowledge. Green text indicates the correct statements supported by the tabular data, and red text indicates the incorrect statements.}
    \label{casestudy}
\end{figure}

\textbf{Case Study}. Figure~\ref{casestudy} shows an output generated by the Bart-base model, providing a deeper insight into how our framework works. The text colored in green indicates facts confirmed by tabular data, while the text colored in red indicates incorrect facts. These results suggest that the model trained with domain-specific knowledge outperforms its counterpart without such knowledge, in particular showing significant improvements in fluency and fidelity. In contrast, the model without domain-specific knowledge tends to hallucinate and generate content that is inconsistent with the original tabular data. However, despite these improvements, both models show inadequate performance in capturing highlighted content.

%% file: Files/conclusion.tex
\section{Conclusion}
We introduce a rigorous task, controlled table-to-text generation with scientiﬁc reasoning, to assess the machine’s ability to produce analytical descriptions that satisfy domainspeciﬁc knowledge and user preferences. To support research in this area, we present a novel dataset, CTRLSciTab, which is characterized by two features: (1) its table-description pairs from the scientiﬁc literature require domain-speciﬁc knowledge for scientiﬁc reasoning, and (2) all descriptions are consistent with user preferences. To evaluate the effectiveness of CTRLSciTab, we employ different baselines and perform comprehensive automatic and human evaluations. Our study reveals key ﬁndings: (1) existing evaluation metrics inadequately measure controlled table-to-text generation with scientiﬁc reasoning; (2) the dominant GPT-3.5 model struggles with tasks requiring advanced expertise; (3) several challenges are associated with the CTRLSciTab dataset. Despite these issues, the initial results lay the groundwork for future research on pre-training tasks for complex, realistic domains. Overall, we position CTRLSciTab as a valuable asset to the research community and expect it to inspire further exploration in this area.

\section{Acknowledgments} 
We thank the anonymous reviewers for their thoughtful comments. This work was partially supported by National Key R\&D Program of China (No.2022YFB3904204), NSF China (No. 42050105, 62272301, 62020106005, 62061146002, 61960206002). This work is supported by the Deep-time Digital Earth (DDE) Big Science Program.